\documentclass[10pt,twocolumn,letterpaper]{article}

\usepackage{times}
\usepackage{epsfig}
\usepackage{graphicx}
\usepackage{amsmath}
\usepackage{amssymb}

\usepackage{bm}
\usepackage{xcolor}
\usepackage[linesnumbered,ruled,vlined]{algorithm2e}
\usepackage{times}
\usepackage{epsfig}
\usepackage{graphicx}
\usepackage{multirow}
\usepackage{amsmath}
\usepackage{amssymb}
\usepackage{verbatim}
\usepackage[linesnumbered,ruled,vlined]{algorithm2e}
\usepackage[acronym,smallcaps]{glossaries}

\usepackage{xr-hyper}
\usepackage[pagebackref=true,breaklinks=true,colorlinks,bookmarks=false]{hyperref}

\newacronym{pc}{PC}{point cloud}
\newacronym{ig}{IG}{Integrated Gradients}
\newacronym{cta}{CTA}{critical traverse attack}
\newacronym{opa}{OPA}{one-point attack}
\newacronym{dnn}{DNN}{deep neural networks}
\newacronym{am}{AM}{Activation Maximization}

\newcommand{\beginsupplement}{%
        \setcounter{table}{0}
        \renewcommand{\thetable}{S\arabic{table}}%
        \setcounter{figure}{0}
        \renewcommand{\thefigure}{S\arabic{figure}}%
        \setcounter{equation}{0}
        \renewcommand{\theequation}{S\arabic{equation}}%
     }

\begin{document}

%%%%%%%%% TITLE
\title{Explainability-Aware One Point Attack for Point Cloud Neural Networks}

\author{Hanxiao Tan\qquad Helena Kotthaus\\
AI Group, TU Dortmund\\
{\tt\small \{hanxiao.tan,helena.kotthaus\}@tu-dortmund.de}
}
\date{}
\maketitle

\thispagestyle{empty}
%%%%%%%%% ABSTRACT
\begin{abstract}
With the proposition of neural networks for point clouds, deep learning has started to shine in the field of 3D object recognition while researchers have shown an increased interest to investigate the reliability of point cloud networks by adversarial attacks. However, most of the existing studies aim to deceive humans or defense algorithms, while the few that address the operation principles of the models themselves remain flawed in terms of critical point selection. In this work, we propose two adversarial methods: One Point Attack (OPA) and Critical Traversal Attack (CTA), which incorporate the explainability technologies and aim to explore the intrinsic operating principle of point cloud networks and their sensitivity against critical points perturbations. Our results show that popular point cloud networks can be deceived with almost $100\%$ success rate by shifting only one point from the input instance. In addition, we show the interesting impact of different point attribution distributions on the adversarial robustness of point cloud networks. Finally, we discuss how our approaches facilitate the explainability study for point cloud networks. To the best of our knowledge, this is the first point-cloud-based adversarial approach concerning explainability. Our code is available at \url{https://github.com/Explain3D/Exp-One-Point-Atk-PC}.
\end{abstract}

%%%%%%%%% BODY TEXT
\section{Introduction}
%Establish the context, background and/or importance of the topic

\begin{comment}
\begin{figure}
    \centering
    \includegraphics{figures/1pexp}
    \caption{One point attack for point cloud networks. With the saliency map provided by the explainability method, only one point needs to be perturbed in the point set of the original instance to fool the most popular point cloud networks.}
    \label{1page_example}
\end{figure}
\end{comment}

Developments in the field of autonomous driving and robotics have heightened the need for the research of \gls{pc} data since \gls{pc}s are advantageous over other 3D representations for real-time performance. However, compared with 2D images, the robustness and reliability of \gls{pc} networks have only attracted considerable attention in recent years and still not been sufficiently studied, which potentially threatens human lives e.g. driverless vehicles with point cloud recognition systems are unreliable unless they are sufficiently stable and transparent. 

Several attempts have been made to investigate the adversarial attacks on \gls{pc} networks, e.g. \cite{lee2020shapeadv} and \cite{xiang2019generating}. The first series of approaches (\textit{shape-perceptible}), represented by \cite{lee2020shapeadv}, although produce geometrically continuous adversarial examples with external generative models, are of limited contribution to the exploration of the classifier itself. Another series (\textit{point-shifting}) represented by \cite{xiang2019generating} have shown a possibility that moving (dropping or adding) points at crucial positions can successfully fool the classifier. Nevertheless, most of such studies have only focused on minimizing perturbation distances for imperceptibility. Conversely, we start from a different perspective by exploring attacks on \gls{pc} networks with a minimal number of perturbed points. Additionally, we argue that existing choices of critical points could be further optimized incorporating explainability approaches.

In comparison to previous studies, our work is motivated by the following reasons:

%\begin{enumerate}[label=(\roman*)]
\textit{Vicinity of decisions}: Minimal perturbation facilitates awareness of the vicinity of the input instances. Although most point-shifting methods also minimize perturbations, they place more emphasis on the perturbation distance and neglect the dimensionality. For high-dimensional decision boundaries, reduction of perturbation dimension is an important way to enhance comprehensibility, which can be regarded as "cutting the input space using very low-dimensional slices" \cite{su2019one}.

\textit{Model operating principle}: Part of the point-shifting methods also deceived the classifier by perturbing critical points, however, we argue that their selection of critical points is flawed. On the other hand, \cite{gupta20203d} demonstrated that feature attributions for \gls{pc} classification networks are extremely sparse, while no work has specifically studied how these attributions are distributed among the critical points as well as their impact on the prediction sensitivity.

\textit{Potential for explainability}: Another possibility of one-dimensional perturbations is explainability. The explainability method called counterfactuals alters the prediction label by perturbing the input features to provide a convincing explanation to the users. Previous researches have documented that humans are more receptive to counterfactuals with sparse-dimensional perturbations \cite{keane2020good,miller1956magical,alvarez2004capacity}. Incorporating part semantics, Our approach has the potential to be extended for generating high-quality counterfactuals. Moreover, ours require only the access of gradients and no additional generative models, and are therefore more intrinsically explainable. 

%\end{enumerate}

Altogether, the contribution of this work can be summarized as follows:
\begin{itemize}
   % \item We present a new explainability perspective of \gls{pc} adversarial attacks inspired by counterfactuals~\cite{byrne2019counterfactuals} for tabular data, with more observable prediction mechanism and decision boundaries through extremely low-dimensional points shifting.
    %\item We propose \gls{opa}, a point cloud-applicable shifting-based attack. With the critical selection module \gls{ig}, \gls{opa} fools popular point cloud networks by perturbing \textbf{one point} with high success rates.
    %\item We propose \gls{cta}, which further improves the success rate, limits the perturbation distances and can be easily extended to targeted attack with the cost of only few additional shifted points.
    \item We propose two explainability method-based adversarial attacks: \gls{opa} and \gls{cta}. Incorporating the attribution from explainable AI, our methods fool the popular \gls{pc} networks with almost $100\%$ success rate. Supported by extensive experiments, a significant margin is established with existing approaches in terms of the perturbation sparsity.
    
    \item We investigate diverse pooling architectures as alternatives to existing point cloud networks, which have an impact on the internal vulnerability against critical points shifting.
    
    \item We discuss the research potential of adversarial attacks from an explainability perspective, and present an application of our methods on facilitating the evaluation of explainability approaches.
    
    %\item We discuss a more persuasive viewpoint of evaluating the robustness of point cloud models against adversarial attacks.
    %\item We discuss the research potential of adversarial attacks from an explainability perspective inspired by the counterfactual explainability methods.
\end{itemize}

The rest of the paper is organized as follows: We introduce the related researches of \gls{pc} attacks in section \ref{sec:relatedwork}, then we detailed our proposed methods in section \ref{sec:methods}. In section \ref{sec:experiments}, we present the visualization of the adversarial examples and demonstrate comparative results with existing studies. In section \ref{sec:disc&concl} we discuss interesting observations derived from experiments with respect to robustness and explainability. We finally summarize our work in section \ref{conclusion}.

%Indicate an issue, problem, or controversy in the field of study

%State the purpose/contribution of the essay or piece of writing

%Provide an overview of the coverage and/or structure of the writing

\section{Related Work} \label{sec:relatedwork}
As the first work~\cite{szegedy2013intriguing} on adversarial examples was presented, an increasing variety of attacks against 2D image neural networks followed~\cite{goodfellow2014explaining,carlini2017towards,kurakin2016adversarial,papernot2017practical,dong2018boosting,moosavi2016deepfool}. However, due to the structural distinctions with \gls{pc} networks (see Supplementary section \ref{RWpcnetworks}), we do not elaborate on the attack methods of image \gls{dnn}s. Relevant information about image adversarial examples refers to~\cite{akhtar2018threat}. It is notable that~\cite{su2019one} investigated one-pixel attack for fooling image \gls{dnn}s and also aimed at exploring the boundary of inputs. Nevertheless, their approach is a black-box attack based on an evolutionary algorithm, which is essentially distinct from ours.

%The research on adversarial examples of \gls{pc}s has come into prominence only in recent years, thus not as numerous researches have been devoted to this field as to 2D images. 
Existing \gls{pc} attacks are generally categorized into two classes: (i) \textit{Shape-perceptible} generation, which generates human-recognizable adversarial examples with consecutive surfaces or meshes via generative models or spacial geometric transformations \cite{zhang20213d,lee2020shapeadv,wen2020geometryaware,li2021pointba,hamdi2020advpc,zhao2020isometry,liu2020adversarial,zhou2020lg}. (ii) \textit{Point-shifting} perturbation, which regularize the distance or dimension of the point-wise shifting via perturbing or gradient-aware white-box algorithms \cite{kim2020minimal,zheng2019pointcloud,wicker2019robustness,xiang2019generating,sun2020adversarial,liu2019extending}. %Instead of addressing on human-level perceptibility and escaping from defense algorithms such as outlier removal, 
Point-wise perturbations, especially gradient-aware attack methodologies, enable more intrinsic explorations of the model properties such as stabilities and decision boundaries. On the other hand, from the perspective of explainability, the majority of generative models contain complex network structures that are inherently unexplainable. Utilizing their output to interpret another model is counter-intuitive.

The conception "\textbf{critical points}" has been discussed by part of the approaches in (ii) \cite{xiang2019generating,kim2020minimal,zheng2019pointcloud,wicker2019robustness} as well as the PointNet proposer \cite{qi2017pointnet}, which forms the skeletons of the input instances in the classification processes. Existing methods extract the critical points by tracing the ones that remain active from the pooling layer, or by observing the gradient-based saliency maps. While such approaches succeed in generating adversarial examples with minor perturbation distances and sparse shift dimensions, we argue that their modules for selecting critical points can be further optimized. Due to the complex structure of the network, it is difficult to determine whether the surviving points from the pooling layer conclusively make significant contributions to the prediction. Besides, saliency maps based on raw gradients are defective~\cite{adebayo2018sanity,sundararajan2016gradients}. The above factors may result in the involvement of fake critical points or omission of real ones during the perturbation process, which severely impairs the performance of the adversarial algorithms.

%Is this section necessary?
%\subsubsection{Explainability for \gls{pc} \gls{dnn}s}
\section{Methods} \label{sec:methods}
\SetCommentSty{mycommfont}
\SetKwInput{KwInput}{Input}                % Set the Input
\SetKwInput{KwOutput}{Output}              % set the Output

In this section, we formulate the adversarial problem in general and introduce the critical points set (subsection \ref{problem definition}). We present our new attack approaches (subsection \ref{method opa and cta}). %including stopping criteria settings (subsection \ref{otherdetail}).

\subsection{Problem Statement} \label{problem definition}

Let $P \in \mathbb{R}^{n \times d}$ denotes the given point cloud instance, $f:P\rightarrow y$ denotes the chosen \gls{pc} neural network and $M(a,b): \mathbb{R}^{n_a \times d} \times \mathbb{R}^{n_b \times d}$ denotes the perturbation matrix between instance $a$ and $b$. The goal of this work is to generate an adversarial examples $P' \in \mathbb{R}^{n' \times d}$ which satisfies:
\begin{equation}
\begin{aligned}\label{formulargeneral}
argmin(&\left |\{m \in M(P,P'))|m\neq 0 \right \}| \\+ &\left \| M(P,P')) \right \|):f(P')\neq f(P)
\end{aligned}
\end{equation}
Note that among the three popular attack methods for \gls{pc} data: point adding ($n'>n$), point detaching ($n'<n$) and point shifting ($n'=n$), this work considers point shifting only.

We address the adversarial task in equation \ref{formulargeneral} as a gradient optimization problem. We minimize the loss on the input \gls{pc} instance while freezing all parameters of the network:
\begin{equation} \label{loss}
    L=\alpha \times Z[f(P)]+\beta \times D(P,P')
\end{equation}
where $\alpha$ indicates the optimization rate, $Z[f(P)]$ indicates the neuron unit corresponding to the prediction $f(P)$ which guaranties the alteration of prediction, $D(P,P')$ represents the quantized imperceptibility between the input $P$ and the adversarial example $P'$ and $\beta$ is the distance penalizing weight. The imperceptibility has two major components, namely the perturbation magnitude and the perturbation sparsity. The perturbation magnitude can be constrained in three ways: Chamfer distance (equation \ref{chamfer}), Hausdorff distance (equation \ref{hausdorff}) or simply Euclidean distance. We ensure perturbation sparsity by simply masking the gradient matrix, and with the help of the saliency map derived by the explainability method we only need to shift those points that contribute positively to the prediction to change the classification results, which are termed as "critical points set".

\textbf{Critical points set}: The concept was first discussed by its proposer~\cite{qi2017pointnet}, which contributes to the features of the max-pooling layer and summarizes the skeleton shape of the input objects. They demonstrated an \textit{upper-bound} construction and proved that corruptions falling between the \textit{critical set} and the \textit{upper-bound} shape pose no impact on the predictions of the model. However, the impairment of shifting those critical points is not sufficiently discussed. Previous adversarial researches studied the model robustness by perturbing or dropping critical points set identified through monitoring the max-pooling layer or accumulating loss of gradients~\cite{xiang2019generating,kim2020minimal,zheng2019pointcloud,wicker2019robustness}. Nevertheless, capturing the output of the max-pooling layer struggles to identify the real critical points set due to the lack of transparency in the high-level structures (e.g., multiple MLPs following the pooling layer), while saliency maps based on raw gradients suffer from saturation~\cite{adebayo2018sanity,sundararajan2016gradients}, both of which severely compromise the filtering of the critical point set. We therefore introduce \gls{ig}~\cite{sundararajan2017axiomatic}, the state-of-the-art gradient-based explainability approach, to further investigate the sensitivity of model robustness to the critical points set. The formulation of \gls{ig} is summarized in equation \ref{IGformular}.

\textbf{Similarity metrics for point cloud data}: Due to the irregularity of \gls{pc}s, Manhattan and Euclidean distance are both no longer applicable when measuring the similarity between \gls{pc} instances. Several previous works introduce Chamfer~\cite{kim2020minimal,zhang2019adversarial,xiang2019generating,lee2020shapeadv,liu2020adversarial,zhou2020lg,zhang20213d} and Hausdorff~\cite{zhou2019dup,kim2020minimal,zhang2019adversarial,xiang2019generating,liu2020adversarial,zhou2020lg} distances to represent the imperceptibility of adversarial examples. The measurements are formulated as: 
\begin{itemize}
\item Bidirectional Chamfer distance
\begin{equation}
\begin{aligned} \label{chamfer}
    D_{c}(P_a,P_b)=& \frac{1}{|P_a|}\sum_{p_m\in P_a}\min_{p_n\in P_b}\left \| p_m-p_n \right \|_2 + \\ & \frac{1}{|P_b|}\sum_{p_n\in P_b}\min_{p_m\in P_a}\left \| p_n-p_m \right \|_2 
\end{aligned}
\end{equation}

\item Bidirectional Hausdorff distance
\begin{equation}
\begin{aligned} \label{hausdorff}
    D_{h}(P_a,P_b)=\max(&\max(\min_{p_n\in P_b}\left \| p_m-p_n \right \|_2), \\ & max(\min_{p_m\in P_a}\left \| p_n-p_m \right \|_2))
\end{aligned}
\end{equation}
\end{itemize}

\subsection{Attack Algorithms} \label{method opa and cta}

%Suppose the adversarial sample $x'$ is composed of the original instance $x$ and the mutation matrix $M$:
%\begin{equation}
    %x'=\left \{ p + m_i|p \in x,m_i \in M \right \}
%\end{equation}
%\Gls{opa} forces the perturbation sparsity to be $1$. 
\textbf{One-Point Attack (\gls{opa})}: Specifically, \gls{opa} (see Supplementary algorithm \ref{algoopa} for pseudo-code) is an extreme of restricting the number of perturbed points, which requires:
\begin{equation}
    \left |\{m \in M(P,P'))|m\neq 0 \right \}|= 1
\end{equation}
We acquire the gradients that minimize the activation unit corresponding to the original prediction, and a saliency map based on the input \gls{pc} instance from the explanation generated by \gls{ig}. We sort the saliency map and select the point with the top-$n$ attribution as the critical points ($n=1$ for \gls{opa}), and mask all points excluding the critical one on the gradients matrix according to its index. Subsequently the critical points are shifted with an iterative optimization process. An optional distance penalty term can be inserted into the optimization objective to regularize the perturbation magnitude and enhance the imperceptibility of the adversarial examples. We choose Adam~\cite{kingma2014adam} as the optimizer, which has been proven to perform better for optimization experiments. The optimization process may stagnate by falling into a local optimum, hence we treat every $25$ steps as a recording period, and the masked Gaussian noise weighted by $W_n$ is introduced into the perturbed points when the average of the target activation at period $k+1$ is greater than at period $k$. For the consideration of time cost, the optimization process is terminated when certain conditions are fulfilled and the attack to the current instance is deemed as a failure.

\textbf{Critical Traversal Attack (\gls{cta})}: Due to the uneven vulnerability of different \gls{pc} instances, heuristically setting a uniform sparsity restriction for the critical points perturbation is challenging. \Gls{cta} (pseudo-code presented in Supplementary algorithm \ref{algocta}) enables the constraint of perturbation sparsity to be adaptive by attempting the minimum number of perturbed points for each instance subject to a successful attack. The idea of \Gls{cta} is starting with the number of perturbed points $n$ as $1$ and increasing by $1$ for each local failure until the prediction is successfully attacked or globally failed. Similarly, we consider the saliency map generated by \gls{ig} as the selection criterion for critical points, and the alternative perturbed points are incremented from top-$1$ to all positively attributed points. Again, for accelerating optimization we also select Adam~\cite{kingma2014adam} as the optimizer. Since most \gls{pc} instances can be successfully attacked by one-point shifting through the introduction of Gaussian noise in the optimization process, we discarded the noise-adding mechanism in \gls{cta} to distinguish the experiment results from \gls{opa}. The aforementioned local failure stands for terminating the current $n$-points attack and starting another $n+1$ round, while the global failure indicates that for the current instance the attack has failed. We detail the stopping criteria for \gls{opa} and \gls{cta} in section \ref{otherdetail}.

\section{Experiments} \label{sec:experiments}
In this section, we present and analyze the results of the proposed attack approaches. We demonstrate quantitative adversarial examples in subsection \ref{quanexam} and scrutinize the qualitative result in subsection \ref{qualicompare}. Our experiments\footnote{Our code is available at \url{https://github.com/Explain3D/Exp-One-Point-Atk-PC}} were primarily conducted on PointNet~\cite{qi2017pointnet}, which in general achieves an overall accuracy of $87.1\%$ for the classification task on ModelNet40~\cite{wu20153d}. Moreover, we also extended our approaches to attack the most popular \gls{pc} network PointNet++~\cite{qi2017pointnet++} and DGCNN~\cite{wang2019dynamic}, which dominate the \gls{pc} classification task with $90.7\%$ and $92.2\%$ accuracies respectively. We choose Modelnet40~\cite{wu20153d} as the experimental dataset, which contains $12311$ CAD models ($9843$ for training and $2468$ for evaluation) from $40$ common categories, and is currently the most widely-applied point cloud classification dataset. We randomly sampled $25$ instances for each class from the test set, and then selected those instances that are correctly predicted by the model as our victim samples. We also validate our methods on ShapeNet \cite{chang2015shapenet} dataset. All attacks performed in this section are non-targeted unless specifically mentioned. In all experiments, we only compare the performances among \textbf{point-shifting} attacks, motivated by exploring the peculiarities of \gls{pc} networks. Though previous shape-perceptible approaches such as \cite{zhao2020isometry,hamdi2020advpc,zhou2020lg,lee2020shapeadv,zhang20213d} also addressed adversarial studies of \gls{pc}s, they were devoted to generate adversarial instances with human-perceptible geometries. Comparison of perturbation distances and dimensions with their works is not relevant, and therefore we do not consider them as competitors.

%Parameter selection describ

\subsection{Adversarial examples visualization} \label{quanexam}
\begin{figure*}
    \begin{centering}
    \includegraphics[width=1\textwidth]{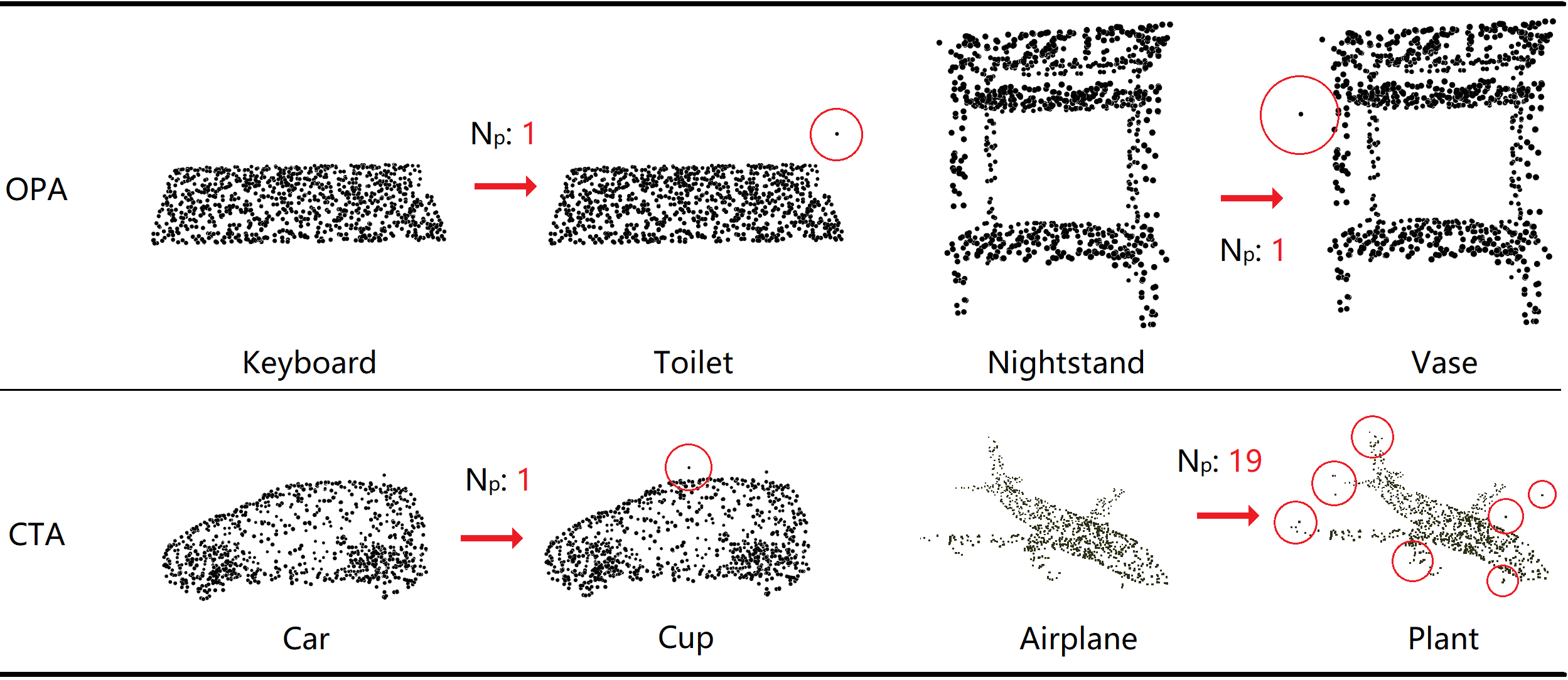}
    \caption{Adversarial examples for \gls{opa} and \gls{cta}. $N_p$ denotes how many points are shifted.}
    \label{visulization1}
    \end{centering}
\end{figure*}

Fig. \ref{visulization1} visualizes two adversarial examples for \gls{opa} and \gls{cta} respectively. Interestingly, in \gls{cta}, regardless of the absence of the restriction on the perturbation dimension, there are instances (e.g. the car in \gls{cta}) where only one-point shifting is required to generate an adversarial example. More qualitative visualizations are presented in Fig. \ref{OPA_100} and \ref{CTA_100}.
 
% Replace A and B
 
\subsection{Quantitative evaluations and comparisons}
\label{qualicompare}

In this section, we compare the imperceptibility of proposed methods with existing attacks via measuring Hausdorff and Chamfer distances as well as the number of shifted points, and demonstrate their transferability among different popular \gls{pc} networks. Additionally, we show that \gls{cta} maintains a remarkably high success rate even after converting to targeted attacks.

\textbf{Imperceptibility}: We compare the quality of generated adversarial examples with other \textbf{point-shifting} researches under the aspect of success rate, Chamfer and Hausdorff distances, and the number of points perturbed. As table \ref{distcompare} shows, compared to the approaches favoring to restrict the perturbation magnitude, despite the relative laxity in controlling the distance between the adversarial examples and the input instances, our methods prevail significantly in terms of the sparsity of the perturbation matrix. Simultaneously, our methods achieve a higher success rate, implying that the network can be fooled for almost all \gls{pc} instances by shifting a minuscule amount of points (even one). In the experiment, the optimization rate $\alpha$ is empirically set to $10^{-6}$, which performs as the most suitable step size for PointNet after grid search. Specifically for \gls{opa}, we set the Gaussian weight $W_n$ to $10^{-1}$, which proved to be the most suitable configuration. More analytical results of different settings of $\beta$ and $W_n$ is demonstrated in Fig. \ref{var_parameters}. Note that while calculating $D_c$ and $D_h$, we employ the L2-norm. Therefore, despite the large Hausdorff distance, the average perturbation magnitude along each axis is $\bm{0.488}$. Considering that each axis of ModelNet40 is regularized into the interval $[-1,1]$, this magnitude occupies $\bm{24.4\%}$ of the interval, which corresponds to an average perturbation of $62.2$ gray values in 2D grayscale images. We thus consider the perturbation magnitude to be acceptable. 

To eliminate potential bias, we also test the proposed attack methods with ShapeNet \cite{chang2015shapenet} dataset. As table \ref{diff_dataset} presents, our approaches perform similarly on the two different datasets, and therefore the vulnerable bias in the data distribution of ModelNet40 can be basically excluded.

\begin{table*}[]
\centering
\begin{tabular}{ccccc}
\hline
                                                               & S                            & $D_c$                                       & $D_h$                                       & $N_p$                     \\ \hline
$L_p$ Norm~\cite{xiang2019generating}          & $85.9$                       & $1.77 \times 10^{-4}$                       & $2.38 \times 10^{-2}$                       & $967$                     \\
Minimal selection~\cite{kim2020minimal}        & $89.4$                       & $1.55 \times 10^{-4}$                       & $1.88 \times 10^{-2}$                       & $36$                      \\
Adversarial sink~\cite{liu2020adversarial}     & $88.3$                       & $7.65 \times 10^{-3}$                       & $1.92 \times 10^{-1}$                       & 1024                      \\
Adversarial stick~\cite{liu2020adversarial}    & $83.7$                       & $4.93 \times 10^{-3}$                       & $1.49 \times 10^{-1}$                       & 210                       \\
Random selection~\cite{wicker2019robustness}   & $55.6$                       & $7.47 \times 10^{-4}$                       & $2.49 \times 10^{-3}$                       & 413                       \\
Critical selection~\cite{wicker2019robustness} & $19.0$                       & \bm{$1.15 \times 10^{-4}$} & $9.39 \times 10^{-3}$                       & 50                        \\
Critical frequency~\cite{zheng2019pointcloud}  & $63.2$                       & $5.72 \times 10^{-4}$                       & $2.50 \times 10^{-3}$                       & 303                       \\
Saliency map/L~\cite{zheng2019pointcloud}      & $56.0$                       & $6.47 \times 10^{-4}$                       & $2.50 \times 10^{-3}$                       & 358                       \\
Saliency map/H~\cite{zheng2019pointcloud}      & $58.4$                       & $7.52 \times 10^{-4}$                       & \bm{$2.48 \times 10^{-3}$} & 424                       \\
Ours (\gls{opa})                              & $98.7$                       & $8.64 \times 10^{-4}$                       & $8.45 \times 10^{-1}$                       & \bm{$1$} \\
Ours ($\gls{cta}_{\beta=0}$)                              & \bm{$100$} & $8.92 \times 10^{-4}$                       & $8.19 \times 10^{-1}$                       & $2$                       \\
Ours ($\gls{cta}_{\beta=1e-3}$)                              & $99.6$ & $7.73 \times 10^{-4}$                       & $6.68 \times 10^{-1}$                       & $6$                       \\ \hline
\end{tabular}
\caption{Comparison of existing point-shifting adversarial generation approaches for PointNet, where $S$, $D_c$, $D_h$ and $N_p$ denote the success rate,  Chamfer and Hausdorff distances and the number of shifted points respectively. Part of the records sourced from~\cite{kim2020minimal}. It is worth noting that we only compare the gradient-based \textbf{point-shifting} competitors.}
\label{distcompare}
\end{table*}

\begin{table}[]
\centering
\resizebox{0.5\textwidth}{10mm} {\begin{tabular}{cccccc}
\hline
                                                                   &Dataset            & S      & $D_c$                  & $D_h$                  & $N_p$ \\ \hline
\multirow{2}{*}{\begin{tabular}[c]{@{}l@{}}OPA\end{tabular}} & ModelNet40 & $98.7$ & $8.64 \times 10^{-4}$ & $8.45 \times 10^{-1}$ & $1$  \\ \cline{2-6} 
                                                                   & ShapeNet   & $95.1$ & $8.39 \times 10^{-4}$ & $8.06 \times 10^{-1}$ & $1$  \\ \hline
\multirow{2}{*}{\begin{tabular}[c]{@{}l@{}}CTA\end{tabular}} & ModelNet40 & $100$ & $8.92 \times 10^{-4}$ & $8.19 \times 10^{-1}$ & $2$  \\ \cline{2-6} 
                                                                   & ShapeNet   & $100$  & $8.91 \times 10^{-4}$ & $7.26 \times 10^{-1}$ & $3$  \\ \hline
\end{tabular}}
\caption{Comparison of attack results with ModelNet40 and ShapeNet dataset.}
\label{diff_dataset}
\end{table}

In addition to PointNet, we also tested the performance of our proposed methods on \gls{pc} networks with different architectures. Table \ref{test other networks} summarize the result of attack PointNet, PointNet++ and DGCNN with both \gls{opa} and \gls{cta} respectively. Surprisingly, these state-of-the-art \gls{pc} networks are vulnerable to the one-point attack with remarkably high success rates. On the other hand, \gls{cta} achieves almost $100\%$ success rate fooling those networks while only a single-digit number of points are shifted. Intuitively, \gls{pc} neural networks appear to be more vulnerable compared to image CNNs (\cite{su2019one} is a roughly comparable study since they also performed one-pixel attack with the highest success rate of $71.66\%$) (see table \ref{table opa 2D} and Fig. \ref{2D_examples} in supplementary for results of \gls{opa}). An opposite conclusion has been drawn by~\cite{xiang2019generating}, they trained the PointNet with 2D data and compared its robustness with 2D CNNs against adversarial images. Nevertheless, we argue that the adversarial examples are generated by attacking a 2D CNN, such attacks may not be aggressive for PointNet, which is specifically designed for \gls{pc}s. %We will further discuss the robustness against attacks in section \ref{discussrobust}.

\begin{table}[]
\centering
\resizebox{0.5\textwidth}{14mm} {\begin{tabular}{cccccc}
\hline
                                                                   & Model & S      & $D_c$                 & $D_h$                 & $N_p$ \\ \hline
\multirow{3}{*}{\begin{tabular}[c]{@{}c@{}}O\\ P\\ A\end{tabular}} & PN     & $98.7$ & $\bm{8.45 \times 10^{-4}}$ & $\bm{8.64 \times 10^{-1}}$ & $1$   \\
                                                                   & PN++   & $\bm{99.1}$ & $1.58 \times 10^{-2}$ & $1.61 \times 10^{1}$  & $1$   \\
                                                                   & DGCNN        & $90.9$ & $1.70 \times 10^{-3}$ & $1.69$                & $1$   \\ \hline
\multirow{3}{*}{\begin{tabular}[c]{@{}c@{}}C\\ T\\ A\end{tabular}} & PN     & $\bm{100}$ & $\bm{8.92 \times 10^{-4}}$ & $\bm{8.19 \times 10^{-1}}$ & $6$   \\
                                                                   & PN++   & $99.5$ & $1.22 \times 10^{-2}$ & $8.90 $               & $6$   \\
                                                                   & DGCNN        & $100$  & $2.13 \times 10^{-3}$ & $1.48$                & $\bm{3}$   \\ \hline
\end{tabular}}
\caption{Comparison of attack results on PN(PointNet), PN++(PointNet++) and DGCNN.}
\label{test other networks}
\end{table}

\begin{figure*} 
    \begin{center}
    \includegraphics[width=0.6\textwidth]{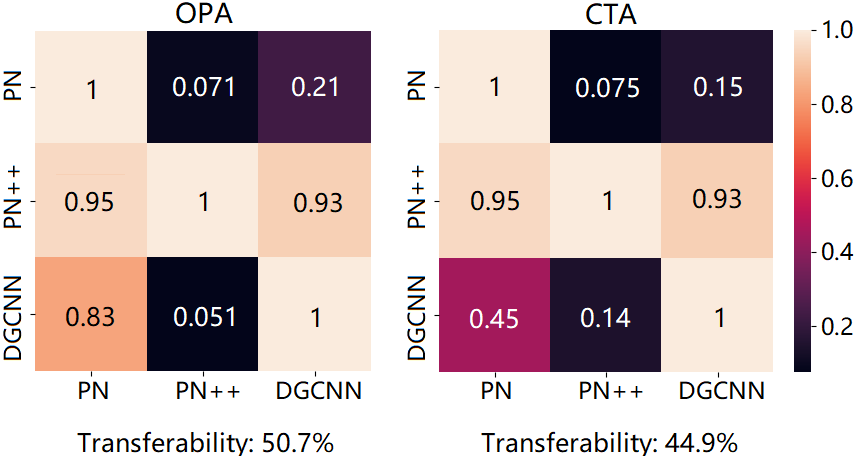}
    \caption{Transferability for PointNet, PointNet++ and DGCNN for \gls{opa} (left) and \gls{cta} (right). Networks on the rows and columns denote from which victim networks the adversarial examples are generated and to which those examples are transferred respectively. Brighter squares denote higher transferabilities. The total transferabilities under the matrices are the averages of the off-diagonal values of corresponding methods.} \label{transferabilityfigure}
    \end{center}
\end{figure*}

\textbf{Transferability}: We further investigate the transferability of proposed attacks across different \gls{pc} networks by feeding the adversarial examples generated by one network to the others and recording the overall classification accuracy. Fig. \ref{transferabilityfigure} presents the adversarial transferability between PointNet, PointNet++ and DGCNN. What stands out in the figure is that PointNet++ and DGCNN show strong stability against the adversarial examples from PointNet. Surprisingly, PointNet++ performs stably against adversarial examples from DGCNN, while the opposite fails. We believe this is because the aggregated adjacency features disperse the attribution of a single point. Recall the \textit{EdgeConv} \cite{wang2019dynamic} in DGCNN, which extracts adjacent features in both point and latent spaces, while PointNet++ possesses a similar module that aggregates neighboring points \cite{qi2017pointnet}, which can be considered as a point-space-only \textit{EdgeConv}. Such an integration distributes the feature contribution to multiple adjacent points, and a modest shifting of one point has limited impacts on the aggregated cluster. The feature extractor in PointNet can also be regarded as a special \textit{EdgeConv} with $K=1$, preserving the location information of the central point only, and therefore is more sensitive to the perturbation. On the other hand, we consider the best stability of PointNet++ stems from the \textit{multi-scale(resolution) grouping}, where latent features are concatenated by grouping layers at different scales, resulting in more points involved in the aggregation. In section \ref{discussrobust}, we perform a preliminary validation of our conjecture on PointNet.

\textbf{Targeted attack}: We also extend the proposed methods to targeted attacks. To alleviate redundant experiment procedures, we employ three alternatives of conducting ergodic targeted attack: \textit{random}, \textit{lowest} and \textit{second-largest} activation attack. In the random activation attack we choose one stochastic target from the $39$ labels (excluding the ground-truth one) as the optimization destination. In the lowest and second-largest activation attack, we decrease the activation of ground truth while boosting the lowest or second-largest activation respectively until it becomes the largest one in the logits. The results, as shown in table \ref{random vs second}, indicate that though the performance of \gls{opa} is deteriorated when converting to targeted attacks due to the rigid restriction on the perturbation dimension, \gls{cta} survived even the worst case (the lowest activation attack) with a remarkably high success rate and a minuscule number of perturbation points. We also demonstrate the results from LG-GAN \cite{zhou2020lg}, which also dedicates to targeted attack for \gls{pc} networks. In comparison, \gls{cta} achieves an approximated success rate with a much smaller $D_c$. Note that their approach is based on generative models and the comparison is for reference only.

\begin{table}
\centering
\resizebox{0.45\textwidth}{14.5mm} {
\begin{tabular}{cccccc}
\hline
                                                                   & Pattern        & S    & $D_c$                 & $D_h$                 & $N_p$ \\ \hline
\multirow{3}{*}{\begin{tabular}[c]{@{}c@{}}O\\ P\\ A\end{tabular}} & Second-largest & 58.5 & $9.49 \times 10^{-4}$ & $9.33 \times 10^{-1}$ & 1     \\
                                                                   & Random         & 20.9 & $1.06 \times 10^{-2}$ & $1.08 \times 10^{1}$  & 1     \\
                                                                   & Lowest         & 6.3  & $4.73 \times 10^{-3}$ & 4.80                  & 1     \\ \hline
\multirow{3}{*}{\begin{tabular}[c]{@{}c@{}}C\\ T\\ A\end{tabular}} & Second-largest & 99.5 & $1.55 \times 10^{-3}$ & $8.14 \times 10^{-1}$ & 5     \\
                                                                   & Random         & 97.7 & $5.75 \times 10^{-3}$ & 2.31                  & 10    \\
                                                                   & Lowest         & 99.0 & $8.52 \times 10^{-3}$ & 3.06                  & 13    \\ \hline
\multicolumn{2}{c}{LG-GAN \cite{zhou2020lg}}                                                          & 98.3 & $3.80 \times 10^{-2}$ & -                     & -     \\ \hline
\end{tabular}}
\caption{Targeted \gls{opa} and \Gls{cta} on PointNet. Targeting all labels for each instance in the test set is time-consuming. Therefore, we generalize it with three substitutes: random, the second-largest and the lowest activation in the logits. We also show the results of LG-GAN as a reference.}
\label{random vs second}
\end{table}

\section{Discussion} \label{sec:disc&concl}
%Beyond proposing an adversarial example generating approach, this work also sets out to shed a light on the properties of popular \gls{pc} networks. 
In this section, we present the relevant properties of \gls{pc} networks in the maximization activation experiment (\ref{discussAM}) as well as our viewpoint concerning the robustness of \gls{pc} networks (\ref{discussrobust}) and discuss the investigative potential of \gls{opa} for \gls{pc} neural networks from the viewpoint of explainability (\ref{discussXAI}).

\subsection{Maximized activation} \label{discussAM}
\Gls{am}, first proposed by \cite{erhan2009visualizing}, sets out to visualize a global explanation of a given network through optimizing the input matrix $x$ while freezing all parameters $\theta$ such that the selected $i^{th}$ activation neuron at $l^{th}$ layer $S_i^l$ is maximized~\cite{nguyen2019understanding}:
\begin{equation}
x^* = \underset{x}{\mathrm{argmax}}\, (a_i^l(\theta,x))
\end{equation}

\begin{figure*}
    \begin{centering}
    \includegraphics[width=0.8\textwidth]{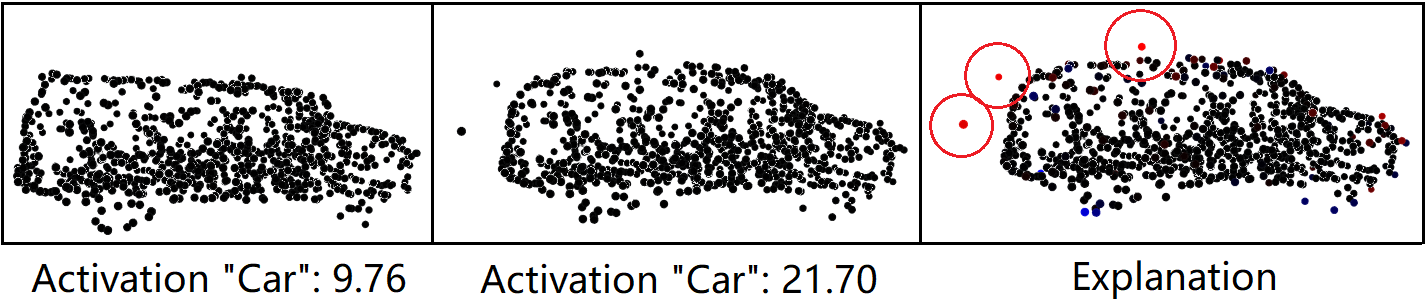}
    \caption{AM results initialized with a certain instance. The first, second and third columns demonstrate the initialized set of points, the AM output results after $1000$ optimization steps and the salience map explanation of the corresponding output explained by IG, respectively. In the explanation, red points indicate the degree of positive attributions.}
    \label{AM}
    \end{centering}
\end{figure*}

\begin{figure*}
    \begin{centering}
    \includegraphics[width=1.0\textwidth]{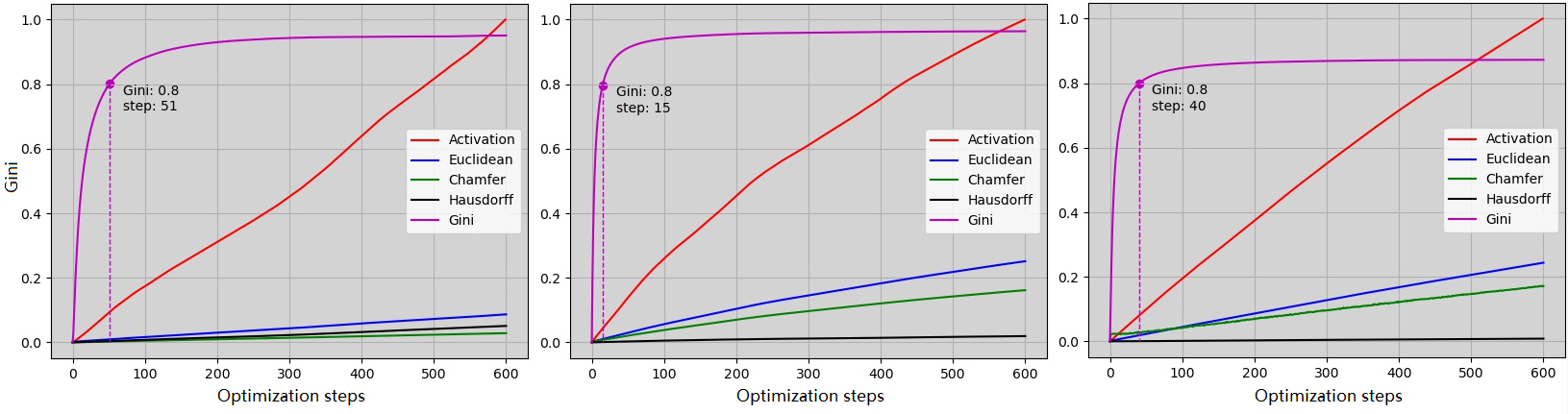}
    \caption{Correlation between the ascending target activation and the various distances of the optimized example from the original initializations: zero (left), the average of the test set (middle) and a certain instance (right). Activations are normalized in order to be visible together with other factors. X-axis denotes the optimization steps and y-axis denotes corresponding values in the legend. The marked points are the steps in the optimization process where the Gini coefficient of the attribution first reaches $0.8$.}
    \label{AMtrend}
    \end{centering}
\end{figure*}

The proposed \gls{opa} was motivated by a fruitless \gls{am} attempt for \gls{pc} networks. Fig. \ref{AM} displays an example from $1000$-steps \gls{am} results of PointNet. More examples with different initializations are depicted in Fig. \ref{AM_zero_avg}. We conduct the \gls{am} experiments with various initializations including zero, point cluster generated by averaging all test data~\cite{nguyen2016multifaceted} and a certain instance from the class "Car".  What stands out in the visualization is that the gradient ascent of the \gls{pc} neural network's activations appears to depend solely on the magnitude of the outward extension subject to the extreme individual points (the middle figure). We further investigate the explanations of the \gls{am} generations utilizing \gls{ig} and the analysis reveals that almost all the positive attributions are concentrated on the minority points that were expanded (the right figure). Fig. \ref{AMtrend} provides a quantitative view of how target activation ascends with the shifting of input points and we introduce Gini coefficient~\cite{dorfman1979formula} to represent the "wealth gap" of the Euclidean distance among all points. Interestingly, as the target activation increments over the optimization process, the Gini coefficient of Euclidean distances steepens to $1$ within few steps, indicating that the fastest upward direction of the target activation gradient corresponds with the extension of a minority of points.
\begin{table*}[h]
\centering
\begin{tabular}{ccccccc}
\hline
                & Acc. & S    & $D_c$                 & $D_h$                 & $N_{pos}$ & Gini.  \\ \hline
Max-pooling     & 87.1 & 98.7 & $8.45 \times 10^{-4}$ & $8.64 \times 10^{-1}$ & 397.2     & $0.91$ \\
Average-pooling & 83.8 & 44.8 & $2.94 \times 10^{-3}$ & $2.96$                & 718.5     & $0.53$ \\
Median-pooling  & 74.5 & 0.9  & $1.28 \times 10^{-4}$ & $9.55 \times 10^{-2}$ & 548.1     & $0.57$ \\
Sum-pooling     & 76.7 & 16.7 & $2.50 \times 10^{-3}$ & $2.53$                & 868.2     & $0.49$ \\ \hline
\end{tabular}
\caption{Model accuracies, success attacking rates, average Chamfer and Hausdorff distances of \gls{opa} on PointNet with max, average, median and sum-pooling on the last layer respectively. The evaluation accuracy is also presented in the second column. $N_{pos}$ denotes how many points are positively attributed to the prediction, and Gini. denotes the Gini coefficient of the corresponding attribution distributions.}
\label{max vs avg}
\end{table*}

\subsection{Structural stability of \gls{pc} networks} \label{discussrobust}
Plenty of researches have discussed defense strategies against intentional attacks for \gls{pc} networks~\cite{zhang2019adversarial,liu2019extending,zhou2019dup,liu2020adversarial,kim2020minimal,sun2020adversarial,zhou2020lg,zhang20213d}, the majority of which were with respect to embedded defense modules, such as outlier removal. However, there has been little discussion about the stability of the intrinsic architectures for \gls{pc} networks. Inspired by \cite{sun2020adversarial} who investigated the impacts of different pooling layers on the robustness, we replace the max-pooling in PointNet with multifarious pooling layers. As table \ref{max vs avg} shows, although PointNet with average and sum-pooling sacrifice $3.3\%$ and $10.4\%$ accuracies in the classification task, the success rates of \gls{opa} on them plummet from $98.7\%$ to $44.8\%$ and $16.7\%$ respectively, and the requested perturbation magnitudes are dramatically increased, which stands for enhanced stabilization. We speculate that it depends on how many points from the input instances the model employs as bases for predictions. We calculate the normalized \gls{ig} contributions of all points from the instances correctly predicted among the $2468$ test instances, and we also introduce the Gini coefficient~\cite{dorfman1979formula} to quantify the dispersion of the absolute attributions which is formulated as:

\begin{equation}
    G = \frac{\sum_{i=1}^{n}\sum_{j=1}^{n}\left | \left |a_i  \right |-\left |a_j  \right | \right |}{2n^2\left |\bar{a}  \right |}
\end{equation}
where $a$ is the attribution mask generated by \gls{ig}.
We demonstrate the corresponding results in table \ref{max vs avg}, \ref{Attribution distribution table} and Fig. \ref{attribution distribution}. There are significant distributional distinctions between the max, average and sum-pooling architectures. PointNet with average and sum-poolings adopt $70.18\%$ ($718.5$ points) and $84.78\%$ ($868.2$ points) of the points to positively sustain the corresponding predictions, where the percentages of points attributed to the top $20\%$ are $0.65\%$ ($6.7$ points) and $1.16\%$ ($11.9$ points), respectively, while these proportions are only $38.79\%$ ($397.2$ points) and $0.15\%$ ($1.5$ points) in the max-pooling structured PointNet. Moreover, the Gini coefficients reveal that in comparison to the more even distribution of attributions in average ($0.53$) and sum-pooling ($0.49$), the dominant majority of attributions in PointNet with max-pooling are concentrated in a minuscule number of points ($0.91$). Hence, it could conceivably be hypothesized that for \gls{pc} networks, involving and apportioning the attribution across more points in prediction would somewhat alleviate the impact of corruption at individual points on decision outcomes, and thus facilitate the robustness of the networks. Surprisingly, median-pooling appears to be an exception. While the success rate of \gls{opa} is as low as $0.9\%$, the generated adversarial examples only require perturbing $9.55 \times 10^{-2}$ of the Hausdorff distance in average (all experiments sharing the same parameters, i.e. without any distance penalty attached). On the other hand, despite that merely $53.53\%$ ($548.1$) points are positively attributed to the corresponding predictions, with only $0.23\%$ ($2.4$ points) of them belonging to the top $20\%$, which is significantly lower than the average and sum-pooling architectures, median-pooling is almost completely immune to the deception of \gls{opa}. We believe that median-pooling is insensitive to extreme values, therefore the stability to perturbations of a single point is dramatically reinforced.

\begin{table}[]
\centering
\begin{tabular}{cccc}
\hline
                & Top $20\%$ & Top $40\%$ & Positive  \\ \hline
Max-pooling     & $0.15\%$   & $0.23\%$   & $38.79\%$ \\
Average-pooling & $0.65\%$   & $2.12\%$   & $70.18\%$ \\
Median-pooling  & $0.23\%$   & $0.59\%$   & $53.53\%$ \\
Sum-pooling     & $1.16\%$   & $4.53\%$   & $84.78\%$ \\ \hline
\end{tabular}
\caption{Overview of the percentage of top-$20\%$, top-$40\%$ and positive attributed points with four different pooling layers. Complete pie diagrams are shown in Fig. \ref{attribution distribution}.}
\label{Attribution distribution table}
\end{table}

\subsection{Towards explainable \gls{pc} models} \label{discussXAI}
Despite the massive number of adversarial methods that have made significant contributions to the studies of model robustness for computer vision tasks, to our best knowledge, none has discussed the explainability of \gls{pc} networks. However, we believe that the adversarial methods can facilitate the explainability of the models to some extent. Recall the roles of counterfactuals in investigating the explainability of models processing tabular data~\cite{byrne2019counterfactuals}. Counterfactuals provide explanations for chosen decisions by describing what changes on the input would lead to an alternative prediction while minimizing the magnitude of the changes to preserve the fidelity, which is identical to the process of generating adversarial examples~\cite{dandl2020multi}. Unfortunately, owing to the multidimensional geometric information that is unacceptable to the human brain, existing image-oriented approaches addressed the counterfactual explanations only at the semantic level~\cite{goyal2019counterfactual,vermeire2020explainable}.

\begin{figure}
    \begin{centering}
    \includegraphics[width=0.475\textwidth]{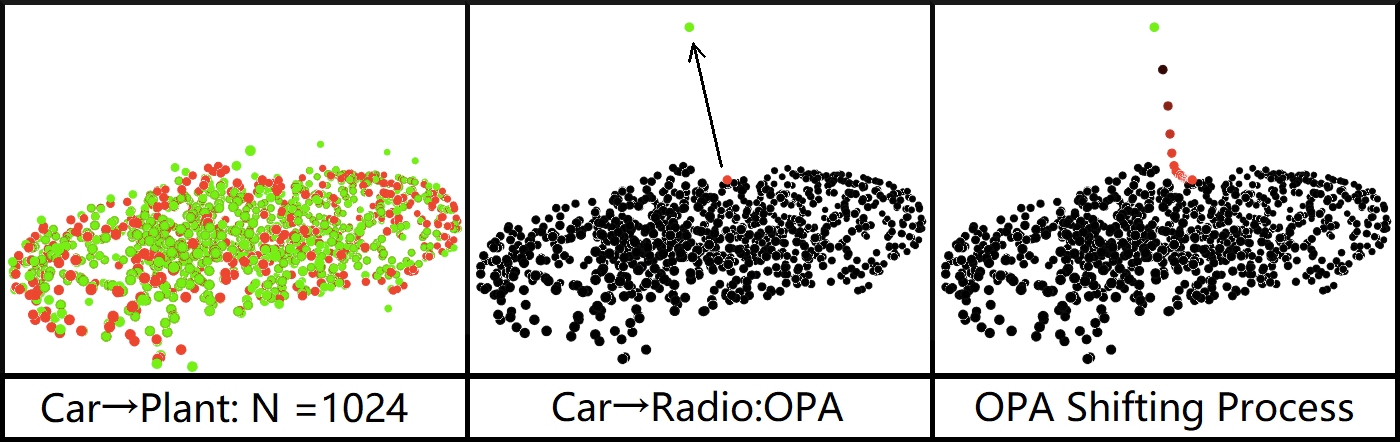}
    \caption{Intuitive visualization of multidimensional shifting(left), unidimensional \gls{opa} shifting(middle) and the shifting process of \gls{opa}. The redder the point the higher the prediction score for the label "car". Green points denote the shifting destinations.}
    \label{visu_optm}
    \end{centering}
\end{figure}

Several studies have documented that a better explanatory counterfactual needs to be sparse because of the limitations on human category learning~\cite{keane2020good} and working memory~\cite{miller1956magical,alvarez2004capacity}. Therefore we argue that unidimensional perturbations contribute to depicting relatively perceptible decision boundaries. Fig. \ref{visu_optm} compares the visualization of multidimensional and unidimensional perturbations. The unidimensional shift, though larger in magnitude, shows more clearly the perturbation process of the prediction from "car" to "radio", and makes it easier to perceive the decision boundary. Conversely, while higher dimensional perturbations perform better on imperceptibility for humans, they are more difficult for understanding the working principles of the model. 

In addition, we find another interesting application of the proposed approaches regarding explainability. Evaluating explanations has long been a major challenge for explainability studies due to the lack of ground truth \cite{burkart2021survey}. An intuitive idea is sensitivity testing, i.e., perturbing features in the explanation that possess high attributions and observing whether the prediction results dramatically change. Theoretically, in our methods, a more accurate explanation induces a more precise selection of critical points, and therefore a higher success rate when perturbing them for generating adversarial examples. Table \ref{XAI_compare} presents the attack performances utilizing gradient-based explainability methods: Vanilla Gradients \cite{simonyan2014deep}, Guided Back-propagation \cite{springenberg2015striving} and \gls{ig} as the critical identifier respectively. Our results are consistent with \cite{gupta20203d} and \cite{tan2022surrogate}, the performance of \gls{ig} is comparatively better than that of Vanilla Gradients and Back-propagation.

\begin{table}[]
\centering
\begin{tabular}{ccccc}
\hline
Exp. Med. & S               & $D_c$                  & $D_h$                  & $N_p$ \\ \hline
VG & $82.5$          & $8.20 \times 10^{-4}$ & $8.01 \times 10^{-1}$ & 1    \\ \hline
GB & $83.4$          & $8.21 \times 10^{-4}$ & $8.02 \times 10^{-1}$ & 1    \\ \hline
IG & $\bm{98.7}$ & $8.64 \times 10^{-4}$ & $8.45 \times 10^{-1}$ & 1    \\ \hline
\end{tabular}
\caption{\gls{opa} performances utilizing various gradient-based explainability methods to identify the critical points, where VG, GB and IG denote Vanilla Gradients \cite{simonyan2014deep}, Guided Back-propagation \cite{springenberg2015striving} and Integrated Gradients respectively.}
\label{XAI_compare}
\end{table}

\section{Conclusion} \label{conclusion}
As the first attack methods for \gls{pc} networks incorporating explainability, though our approaches are easily filtered by defense modules (such as outlier removal, see section \ref{ethical problem}) due to the relatively large perturbation distances, we demonstrate the significance of individual critical points for \gls{pc} network predictions. We also discussed our viewpoints to the robustness of \gls{pc} networks as well as their explainability. In future investigations, it might be possible to distill existing \gls{pc} networks according to the critical points into more explainable architectures. Besides, we are looking forward to higher-quality and human-understandable explanations for \gls{pc} networks.

\clearpage
{\small
\bibliographystyle{ieee_fullname}
\bibliography{egbib}
}
% This version of CVPR template is provided by Ming-Ming Cheng.
\clearpage
\section{Supplementary Material}
This section is a supplement for the main part of the paper. In this section, we detail additional formulas for the backgrounds (\ref{bg}), demonstrate our Pseudo-codes and stopping criteria (\ref{other_implementations}), show more adversarial examples for both \gls{opa} and \gls{cta} respectively (\ref{more adv examples}), visualize the diversity of attacking labels (\ref{label diversity}), discuss the most appropriate hyper-parameter settings (\ref{hyperpara}). We also present the attack result \gls{opa} on 2D images as a comparable reference (\ref{opa for 2D}). Finally, we provide more visualisations of the Activation Maximization (AM) and the attribution distribution of \gls{pc} networks (\ref{More AM results} and \ref{visualize attri distri} respectively.)

\beginsupplement
\subsection{Background} \label{bg}
\subsubsection{Point cloud deep neural networks} \label{RWpcnetworks}
A \gls{pc} input can be represented as $P=\left \{p_0,...,p_n\right \}$, where $p_i\in \mathbb{R}^3$ and $n$ is the number of component points. Compared with 2D images, the structural peculiarity of \gls{pc} data lies in the irregularity: let $R(S)$ be a function that randomly disrupts the order of the sequence $S$, a \gls{pc} classifier $f$ must possess such properties: $f(P)=f(R(P))$, which is regarded as a "symmetric system". The pioneer of \gls{pc} networks is proposed by \cite{qi2017pointnet}, succeeded by employing an symmetric function $g(S)$ and an element-wise transformer $h(p)$ where $f(P)\approx g(\left \{h(p_0),...,h(p_n) \right \})$ (in their experiments a max-pooling is choosen as $g(S)$). PointNet++~\cite{qi2017pointnet++}, the successor of PointNet, further coalesced hierarchical structures by introducing spatial adjacency via grouping of nearest-neighbors. DGCNN~\cite{wang2019dynamic} extended the the predecessors by dynamically incorporating graph relationships between multiple layers. All of the point-based methods achieve satisfactory accuracies on acknowledged \gls{pc} dataset such as ModelNet40~\cite{wu20153d}.

\subsubsection{Integrated Gradients} \label{introIG}
Gradients-based explainability methods are oriented on generating saliency maps of inputs by calculating gradients during propagation. While vanilla gradients severely suffer from attribution saturation~\cite{sundararajan2016gradients}, ~\cite{sundararajan2017axiomatic} proposes \gls{ig} which accumulates attributions from an appropriate baseline before the gradients reach the saturation threshold. \Gls{ig} is formulated as:

\begin{equation} \label{IGformular}
    IG_i=(x_i-x_i')\cdot \int_{\alpha=0}^{1}\frac{\partial F(x'+\alpha (x-x'))}{\partial x} d\alpha 
\end{equation}
Where $x'$ denotes the given baseline.

\subsubsection{Targeted vs. Untargeted attack}
For a given classifier $f$ and its logits $a$, an \gls{pc} input instance $P$ and an adversarial perturbation $A_p$:
\begin{itemize}
    \item \textit{Targeted attack}
    \begin{equation} \label{untargeted}
        \textrm{Minimize}\quad(a[f(P)+A_p])\quad\textrm{s.t.}\quad f(P+A_p) \neq f(P)
    \end{equation}
    
    \item \textit{Untargeted attack}
    \begin{equation} \label{targeted}
        \textrm{Maximize}\quad(a[f(P+A_p)])\quad\textrm{s.t.}\quad f(P+A_p) = T
    \end{equation}
    Where $T$ is the given target class.
\end{itemize}

\subsection{Other implementation details} \label{other_implementations}
\subsubsection{Pseudo-codes of \gls{opa} and \gls{cta}} \label{pseudocodes}
In this section we present the Pseudo-codes for both \gls{opa} and \gls{cta} as a supplement for section \ref{method opa and cta}.

\subsubsection{Stopping Criteria} \label{otherdetail}
Theoretically, \gls{cta} can keep searching until all positively contributed points are traversed. For algorithmic efficiency, we set specific stopping criteria for \gls{opa} and \gls{cta}.

\textbf{\gls{opa}}: With the introduction of Gaussian random noise for \gls{opa}, the optimization process may fall into an everlasting convergence-noise addition loop, a manually configured failure threshold is therefore essential. A recorder $R_a$ is built to record the corresponding prediction activation for each period. We set a global maximum iterations $I_{maxg}$. The stopping criterion of \gls{opa} is fulfilled when
\begin{itemize}
    \item $I_{cur} > I_{maxg}$ \textbf{or} $((Mean(R_a^{k+1})>Mean(R_a^{k})$ \textbf{and} $Var(R_a^{k})\rightarrow 0))$.
\end{itemize}
Due to the introduction of random Gaussian noise, the optimization process will not fail until the target activation has no fluctuant reaction to the Gaussian noise.

\textbf{\gls{cta}}: There are both local and global stopping criteria for \gls{cta}. Local criterion stands for terminating the current $N_p$ perturbed points and start the $N_p+1$ round, which is similar with \gls{opa}. Again, we set an activation recorder $R_a$ and a \textbf{local} maximum iterations $I_{maxl}$. The local stopping criterion is fulfilled when:
\begin{itemize}
    \item $I_{cur}>I_{maxl}$ \textbf{or} $Mean(R_a^{k+1})>Mean(R_a^{k})$
\end{itemize}
Global stopping terminates the optimization of the current instance and registers it as "failed". \gls{cta} is designed to shift all the positively attributed points $N_{pos}$ in the worst case which is extremely time-consuming. For practical feasibility, we specify the \textbf{global} maximum iterations $I_{maxg}$. The global stopping criterion for \gls{cta} is fulfilled when:
\begin{itemize}
    \item $I_{cur}>I_{maxg}$ \textbf{or} $N_p\geqslant N_{pos}$
\end{itemize}
where $N_{pos}$ is the total amount of positive attributed points according to the explanation provided by \gls{ig}.

\begin{algorithm}
\DontPrintSemicolon
  \KwInput{$P \rightarrow N \times D$ \gls{pc} data, $f \rightarrow$ \Gls{pc} neural network, $\alpha \rightarrow$ Optimizing rate, $\beta \rightarrow$ Weight for constrain the perturbing distance(optional), $D\rightarrow$ Distance calculating function(optional), $N_p \rightarrow$ Number of shifting points($1$ for \textit{One-point attack}), $W_n\rightarrow$ Gaussian noise weights}
  \KwOutput{$P_{adv} \rightarrow N \times D$ Adversarial example}
  
  $A_{idx}=Argsort(IG(P,f))$ \tcp*{Get \gls{ig} mask of P}
  
  $R_{s} = list()$ \tcp*{Activation Recorder}
  
  $I_{cur} = 1$ \tcp*{Current iteration}
  
  \While{True}
   {
   		$a_p \leftarrow f(P)$ \tcp*{Current activation of predicted class}
   		
   		$G=\alpha*A_p+\beta*D(P_{adv},P)$ \tcp*{Add distance constrain(Optional)}
   		
   		$P_{adv}=Adam(P_{adv},G[A_{idx}[1:N]])$ \tcp*{Adam optimizing N points}
   		
   		$I_{cur}\mathrel{{+}{=}}1$
   		
   		$R_{s}.append(a_p)$
   		
   		\tcc{Add masked Gaussian random noise if activation descending stopped}
   		\If{$R_{s}[t] < R_{s}[t+1]$ }
            {
                $P_{adv}\mathrel{{+}{=}}W_n\times GaussianRandom(P_{\delta})[A_{idx}[1:N]]$
            }
        \tcc{Success if predict class changed}
        \If{$max(a)!=pred$}
            {
                return $P_{adv}$
            }
            
        \tcc{Fails if the stopping conditions related to $R_a$ and $I_{cur}$ are fulfilled}
        \If{Stopping criteria are fulfilled}
        {
            return \textit{Failed}
        }
   }
\caption{N-critical Point(s) Attack. ($n=1$ for \gls{opa})}\label{algoopa}
\end{algorithm}

%%%%%%%%%%%%%%%%%%%%%%%%%%%%%%%%%%%%%%%%%%%%%%%%%%%%%%
\begin{algorithm}
\DontPrintSemicolon
  
  \KwInput{$P \rightarrow N \times D$ \gls{pc} data, $f \rightarrow$ \Gls{pc} neural network, $\alpha \rightarrow$ Optimizing rate, $\beta \rightarrow$ Weight for constrain the perturbing distance(optional), $D\rightarrow$ Distance measuring function(optional)}
  \KwOutput{$P_{adv} \rightarrow N \times D$ Adversarial example}
  $A_{idx}=Argsort(IG(P,f))$ \tcp*{Get \gls{ig} mask of P}
  
  $Num_{pos}=count(IG(P,f)>0)$ \tcp*{\# Points with attribution \textgreater 0}
  
  $R_{s} = list()$ 
  
  $I_{cur} = 1$
  
  \For{$N_p$ from $1$ to $Num_{pos}$}
  {
      \While{True}
       {
       		$a_p \leftarrow f(P)$ \tcp*{Activation of predicted class}
       		
            $G=\alpha*A_p+\beta*D(P_{adv},P)$ \tcp*{Add distance constrain(Optional)}
       		
       		$P_{adv}=Adam(P_{adv},G[A_{idx}[1:N_p]])$ \tcp*{Adam optimizing N points}
       		
       		$I_{cur}\mathrel{{+}{=}}1$
       		
       		$R_{s}.append(a_p)$
       		
            \tcc{Success if predict class changed}
            \If{$argmax(a_p)!=pred$}
                {
                    return $P_{adv}$
                }
                
            \tcc{Current $N_p$ round fails if the local stopping conditions related to $R_a$ and $I_{cur}$ are fulfilled}
            \If{\textbf{Local stopping criteria} fulfilled}
            {
                break;
            }
       }
       \tcc{Current instance fails if the global stopping conditions are fulfilled}
    \If{\textbf{Global stopping criteria} fulfilled}
    {
        return \textit{Failed}
    }
    return \textit{Failed}
    }
\caption{Critical Traversal Attack (\gls{cta})}\label{algocta}
\end{algorithm}

\subsection{More qualitative visualizations for \gls{opa} and \gls{cta}} \label{more adv examples}

\begin{figure*}
    \begin{centering}
    \includegraphics[width=\textwidth,height=0.75\textheight]{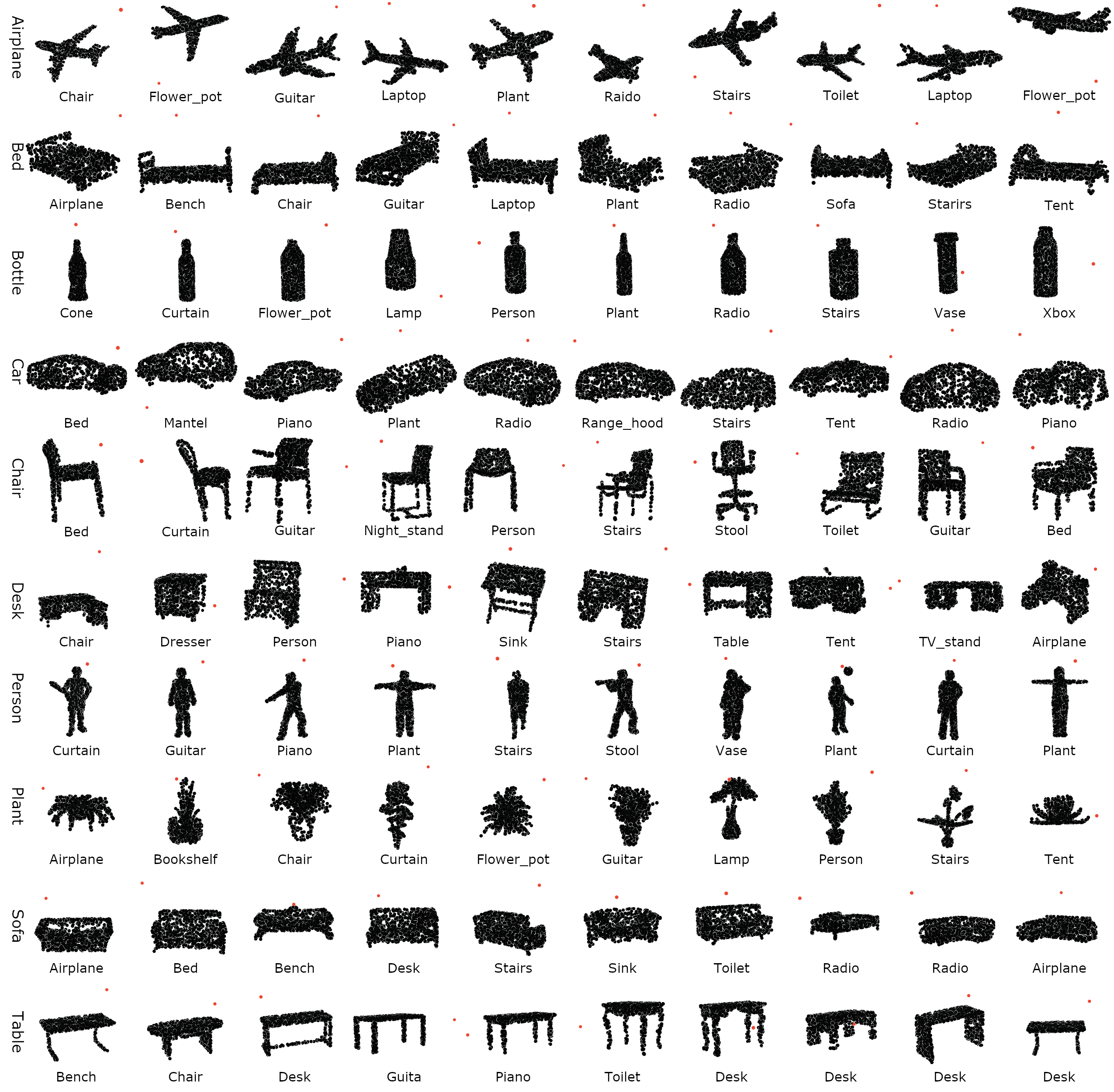}
    \caption{More results from \gls{opa}. We chose the $10$ representative classes that appear more frequently in the real world. The perturbed points are indicated in red to be noticeable.}
    \label{OPA_100}
    \end{centering}
\end{figure*}

\begin{figure*}
    \begin{centering}
    \includegraphics[width=\textwidth,height=0.75\textheight]{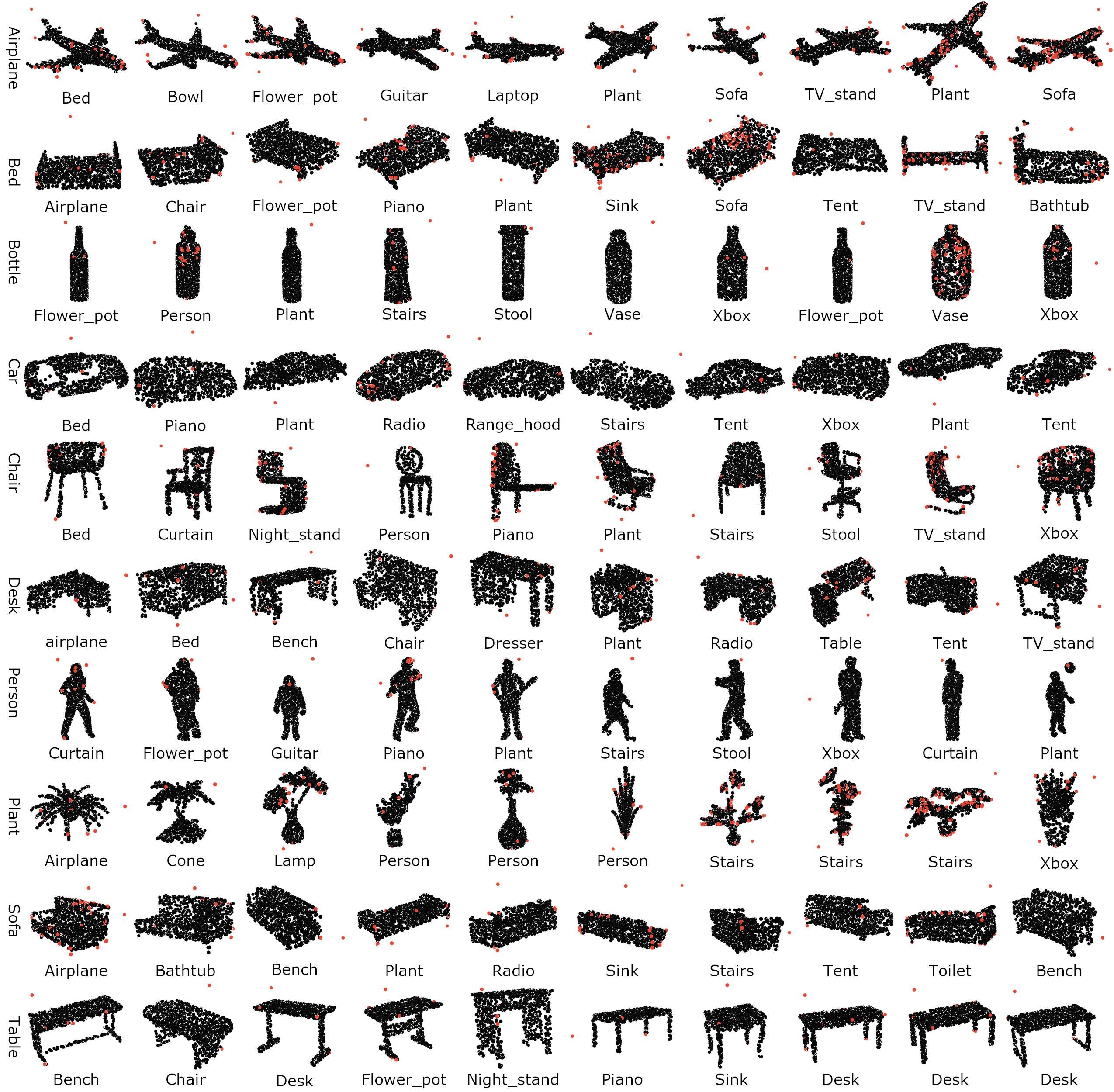}
    \caption{More results from \gls{cta}. We also chose the $10$ representative classes that appear more frequently in the real world. The perturbed points are indicated in red to be noticeable.}
    \label{CTA_100}
    \end{centering}
\end{figure*}

We selected $10$ representative classes from Modelnet40 that occur most frequently in the real world and demonstrate another $10$ adversarial examples for each class generated by \gls{opa} and \gls{cta} in Fig. \ref{OPA_100} and \ref{CTA_100} respectively. The perturbed points are colored with red for better visualization. As the success rate of the \gls{opa} attack is close to $100\%$, in order to distinguish the results of \gls{cta} from \gls{opa} more clearly, we set $\beta$ in \gls{cta} as $(8\times \alpha)$. This setting makes a good trade-off between success rate, shifting distance and perturbation dimensionality. The detailed experimental results are demonstrated in section \ref{hyperpara}.

\subsection{Label Diversity of adversarial examples} \label{label diversity}
\begin{figure*}
    \begin{centering}
    \includegraphics[width=1\textwidth,height=0.8\textheight]{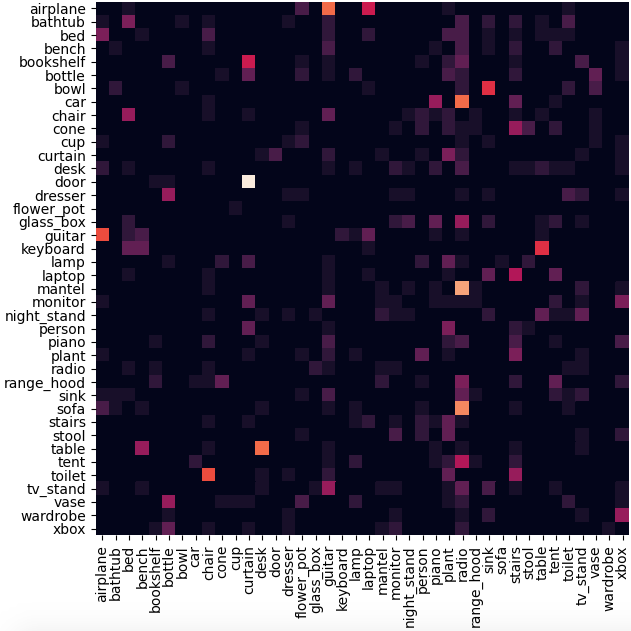}
    \caption{Heat map of successful attacks by \gls{opa} across labels. Rows indicate from which category the adversarial examples come and the columns indicate to which category they are predicted. The brighter the square, the more examples that fall into the corresponding category.}
    \label{opa_classes}
    \end{centering}
\end{figure*}

\begin{figure*}
    \begin{centering}
    \includegraphics[width=1\textwidth,height=0.8\textheight]{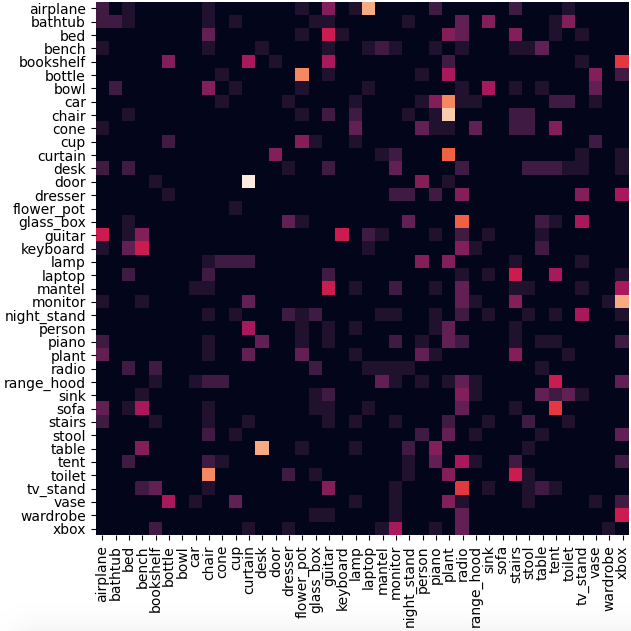}
    \caption{Heat map of successful attacks by \gls{cta} across labels. Rows indicate from which category the adversarial examples come and the columns indicate to which category they are predicted. The brighter the square, the more examples that fall into the corresponding category.}
    \label{cta_classes}
    \end{centering}
\end{figure*}

For non-targeted \gls{opa} and \gls{cta}, the optimization process diminishes the neurons corresponding to the original labels, with no interest in the predicted labels of the adversarial examples. However, we found that observing the adversarial labels helped to understand the particularities of the adversarial examples. Fig. \ref{opa_classes} and \ref{cta_classes} report the label distribution matrices of untargeted \gls{opa} and \gls{cta} respectively. As can be seen from Fig. \ref{opa_classes}, class "radio" is most likely to be the adversarial label, and most of the adversarial examples generated within the same class are concentrated in one of the other categories (e.g. almost all instances from "door" are optimized towards "curtain"). This phenomenon is significantly ameliorated in \gls{cta} (see Fig. \ref{cta_classes}). The target labels are more evenly distributed in the target label matrix, yielding more diversity in the adversarial examples.

\subsection{Hyper-parameter settings} \label{hyperpara}
\subsubsection{Distance regularization $\beta$}
\begin{figure*}
    \begin{centering}
    \includegraphics[width=1\textwidth,height=0.2\textheight]{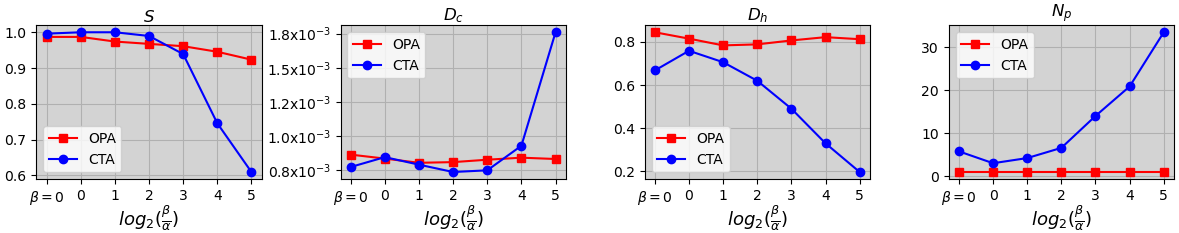}
    \caption{Performance (success rate, Chamfer and Hausdorff distances and the number of shifted points respectively) of \gls{opa} and \gls{cta} in different settings of hyper-parameters. The x-axis indicates the logarithm of the quotient of $\beta$ and $\alpha$ where the first tick denotes $\beta=0$.}
    \label{var_parameters}
    \end{centering}
\end{figure*}

For both proposed algorithms, there are two crucial hyper-parameters to be tuned that affect the performance of the attacks, i.e. $\alpha$ and $\beta$. $\alpha$ indicates the optimization rate and is empirically set to $1e-6$. $\beta$ indicates the penalty of perturbation distances, which regularizes the shifting magnitude and preserves the imperceptibility of adversarial examples. In previous experiments, we temporarily set $\beta$ to $0$ to highlight the sparse perturbation dimensions. However, additional investigations suggest that appropriate $beta$ can further improve the performance of the proposed approaches. Fig. \ref{var_parameters} demonstrates the performances with different $\beta$ settings. Interestingly, we found that \gls{cta} performs best when $\beta=\alpha$: while maintaining nearly $100\%$ success rate and comparably shifting distances, its average $N_p$ dramatically decreases to $3.04$ (different from \gls{opa}, \gls{cta} employs no random-noise). We strongly recommend restricting $\beta$ to a reasonable range ($\leq (8\times \alpha)$) since large $\beta$ easily leads to an explosion in processing time.

\subsubsection{Gaussian noise weight $W_n$ for \gls{opa}}
\begin{figure*}
    \begin{centering}
    \includegraphics[width=\textwidth,height=0.23\textheight]{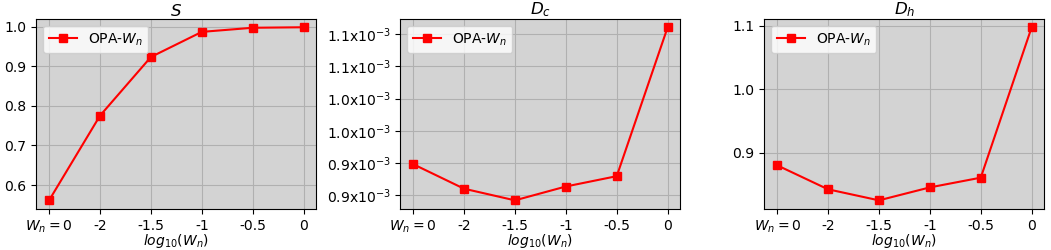}
    \caption{Performance (success rate, Chamfer and Hausdorff distances respectively) of \gls{opa} in different settings of weights for Gaussian noise. The x-axis indicates the logarithm of $W_n$ where the first tick denotes $W_n=0$.}
    \label{opa_noise_par}
    \end{centering}
\end{figure*}

In particular for \gls{opa}, another hyperparameter $W_n$ is set to prevent the optimization process from stagnating at a local optimum. We experimented with various settings of $W_n$ and present the results in Fig. \ref{opa_noise_par}. What stands out in the figure is that the appropriate range for $W_n$ is around $10^{-1}$ to $10^{-0.5}$ where the success rate approximates $100\%$ while maintaining acceptable perturbation distances. Adding Gaussian noise in the optimization process dramatically enhances the attack performance of \gls{opa}, with its success rate increasing from $56.1\%$ as a simple-gradient attack to almost $100\%$. Interestingly, we observe that a suitable noise weight concurrently reduces the perturbation distance and thus augments the imperceptibility of the adversarial examples. We attribute this to the promotion of Gaussian noise that facilitates the optimizer to escape from saddle planes or local optimums faster, reducing the number of total iterations. However, overweighting deviates the critical point from the original optimization path, which is equivalent to resetting another starting position in 3D space and forcing the optimizer to start iterating again. While there remains a high probability of finding an adversarial example, its imperceptibility is severely impaired.

\subsection{\gls{opa} on 2D image neural network} \label{opa for 2D}
\begin{figure}
    \begin{centering}
    \includegraphics[width=0.45\textwidth]{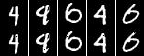}
    \caption{Successful attack examples of ResNet18-GR by \gls{opa}. The first and second rows are input images and adversarial examples respectively.}
    \label{2D_examples}
    \end{centering}
\end{figure}

For a relatively fair comparison as a reference, we extend our \gls{opa} to 2D image neural networks for a rough comparison of its sensitivity to critical points with that of 3D networks. We trained a simple ResNet18 network with the MNIST handwriting dataset, which achieves an accuracy of $99\%$ on the test set. We select $1000$ samples from the test set as victims to be attacked with \gls{opa}. The quantitative results and parts of the adversarial examples are demonstrated in table \ref{table opa 2D} and Fig. \ref{2D_examples} respectively. In Fig. \ref{2D_examples}, the original instances and their adversarial results are listed on the first and the second row respectively. With the removal of a pixel in a critical location, a small number of test images successfully fooled the neural network. However, from a quantitative viewpoint (table \ref{table opa 2D}), shifting one critical point almost fails to fool the ResNet18 network ($1.2\%$ success rate for ResNet18-GR). We believe the reasons are: (1) 2D images are restricted within the RGB/greyscale space, thus there exists an upper bound on the magnitude of the perturbation, while 3D point clouds are infinitely extendable; (2) Large-size convolutional kernels ($\geq 2$) learn local features of multiple pixels, which mitigates the impact of individual points on the overall prediction. According to observation (1), we temporarily remove the physical limitation during attacks to investigate the pure mechanism inside both networks and report the results in ResNet18-GF of table \ref{table opa 2D}. Though the attack success rate climbs to $51.7\%$, there is still a gap with PointNet ($98.7\%$). PointNet encodes points with $1\times 1$ convolutional kernels, which is analogous to an independent weighting process for each point. The network inclines to assign a large weight to individual points due to the weak local correlation of adjacent points and therefore leads to vulnerable robustness against perturbations of critical points.

\begin{table}[]
\centering
\begin{tabular}{cccc}
\hline
            & S      & $D_c$                 & $D_h$                 \\ \hline
ResNet18-GR & $1.2$  & $4.93 \times 10^{-2}$ & $8.67 \times 10^{-1}$ \\
ResNet18-GF & $51.7$ & $1.48$                & $4.08 \times 10^{1}$  \\
PointNet    & $98.7$ & $8.64 \times 10^{-4}$ & $8.45 \times 10^{-1}$  \\ \hline
\end{tabular}
\caption{\gls{opa} attack performance comparisons between ResNet18 and PointNet. ResNet18-GR indicates the attack within the range restriction of the greyscale value ($0\sim 255$), while ResNet18-GF indicates a purely numerical attack possibly with no physical significance (greyscale value less than $0$ or greater than $255$).}
\label{table opa 2D}
\end{table}

\subsection{Additional Activation Maximization (AM) results} \label{More AM results}
\begin{figure*}
    \begin{centering}
    \includegraphics[width=0.7\textwidth]{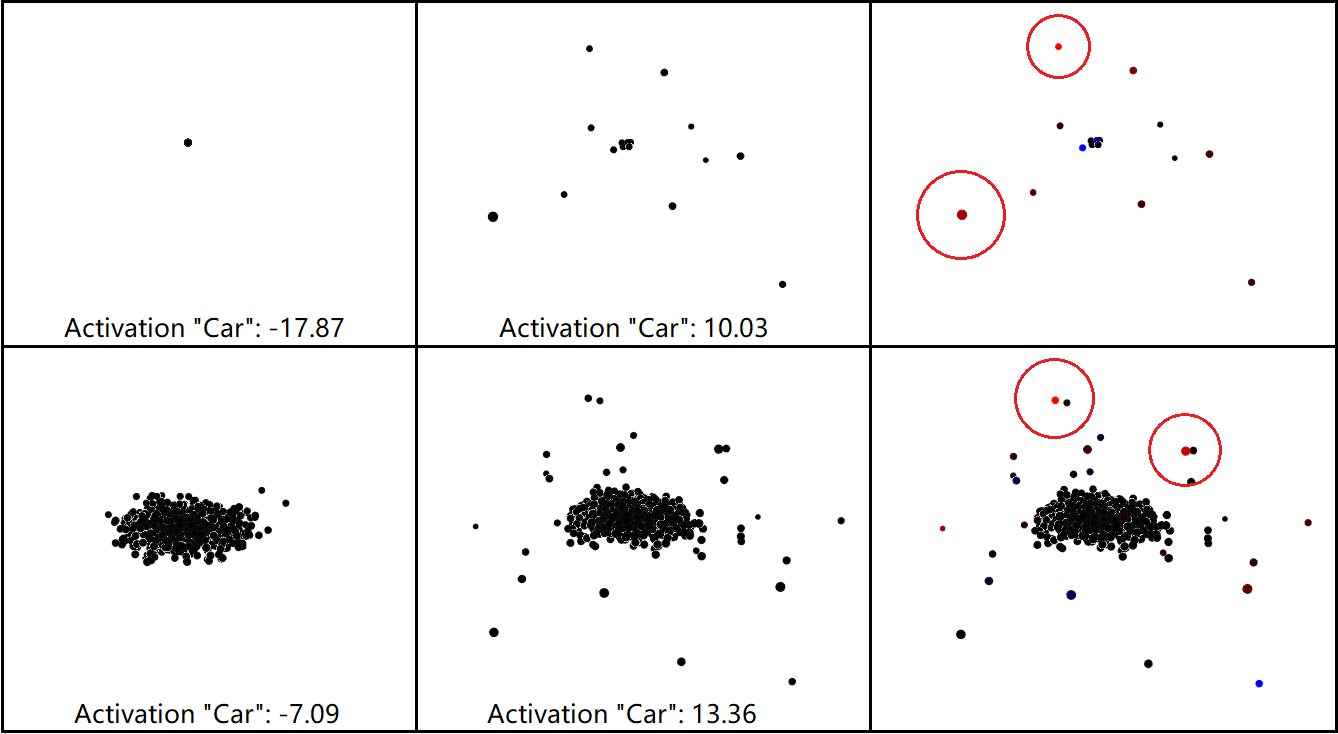}
    \caption{AM results initialized with zeros (the first row) and the point cluster generated by averaging all test data (the second row) respectively. The first, second and third columns demonstrate the initialized set of points, the AM output results after $1000$ optimization steps and the salience map explanation of the corresponding output explained by IG, respectively. In the explanation, red points indicate the degree of positive attributions.}
    \label{AM_zero_avg}
    \end{centering}
\end{figure*}

For fairness and persuasion, we conduct \gls{am} experiments with various initializations as a supplement of section \ref{discussAM}. Fig. \ref{AM_zero_avg} shows AM initialized with zeros and the point cluster generated by averaging all test data~\cite{nguyen2016multifaceted}.

\subsection{Visualization of the attribution distributions} \label{visualize attri distri}
\begin{figure*}
    \begin{centering}
    \includegraphics[width=0.8\textwidth]{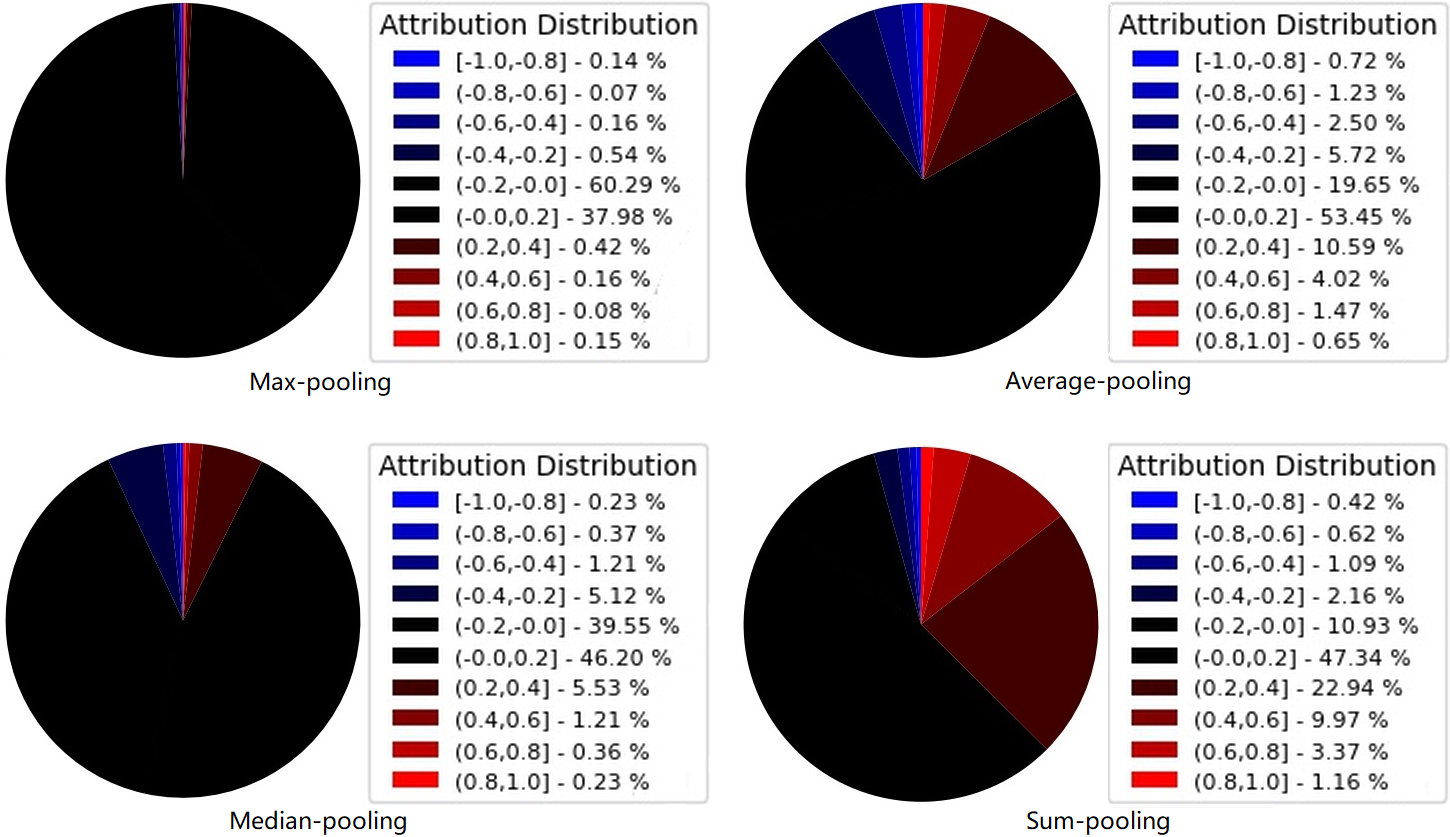}
    \caption{The distributions of attributed points of PointNet structured with max, average, median and sum-pooling layers as the global feature extraction layer respectively.}
    \label{attribution distribution}
    \end{centering}
\end{figure*}

As a supplementary of table \ref{Attribution distribution table}, we demonstrate the complete pie diagrams of the attribution distributions of the aforementioned four pooling structures in \ref{attribution distribution}.

\subsection{Societal impacts and ethical issues} \label{ethical problem}
This work proposes two adversarial approaches, which pose a potential threat to the security of \gls{pc} networks. Motivationally, however, this paper aims to illuminate the distribution of attributions of \gls{pc} networks rather than specifically targeting the attack method of the model. Practically, our proposed approaches can be more easily defended visually or algorithmically compared to related studies aiming at imperceptibility. Table \ref{Sup:table_defence} presents the results of defending against the proposed attacks by a simple outlier removal algorithm. Adversarial samples generated by OPA and CTA are detected with almost $100\%$ success rate. We thus argue that the proposed attacks do not pose a serious threat to existing networks.

\begin{table}[]
\centering
\begin{tabular}{lll}
\hline
    & $r_D$ & $r_P$ \\ \hline
OPA & 98.6 & 100  \\
CTA & 99.2 & 45.6 \\ \hline
\end{tabular}
\caption{Success rates of defense against the proposed attacks by outlier removal. $r_D$ denotes the success rate of the input instance being detected as adversarial examples and $r_P$ denotes the percentage of perturbation points detected by the defense module correctly.}
\label{Sup:table_defence}
\end{table}
\end{document}